# Towards Pedestrian Detection Using RetinaNet in ECCV 2018 Wider Pedestrian Detection Challenge

Md Ashraful Alam Milton
Department of Computer Engineering, Autonomous University of Barcelona
Campus de la UAB, Plaça Cívica, 08193 Bellaterra, Barcelona, Spain
mdashrafulalam.milton@e-campus.uab.cat

## Abstract

*The main essence of this paper is to investigate the performance of RetinaNet based object detectors on pedestrian detection. Pedestrian detection is an important research topic as it provides a baseline for general object detection and has a great number of practical applications like autonomous car, robotics and Security camera. Though extensive research has made huge progress in pedestrian detection, there are still many issues and open for more research and improvement. Recent deep learning based methods have shown state-of-the-art performance in computer vision tasks such as image classification, object detection, and segmentation. Wider pedestrian detection challenge aims at finding improve solutions for pedestrian detection problem. In this paper, We propose a pedestrian detection system based on RetinaNet. Our solution has scored 0.4061 mAP. The code is available at https://github.com/miltonbd/ECCV_2018_pedestrian_detection_challenege .*

1. Introduction

Object detection is a fundamental problem in computer vision which goal is to detect the object class and bounding box location in videos or digital images. Object detection is massively utilized in practical applications such as image retrieval, person identification, video surveillance, etc. Object detection is an integral part of human cognition. We human seamlessly can identify an object's class and location. The artificial intelligent machine must have the perception of the world around it. Pedestrian detection is a branch of object detection problem which deals with detecting the specific human class.Pedestrian detection is a benchmark problem in object detection which has practical application in autonomous driving, surveillance, person identification, face recognition and navigation. Autonomous Driving systems are becoming a reality due to immense progress in computer vision. The highly efficient algorithm and low-cost onboard hardware. However, this problem is still open for research due to the variation of the pedestrian detection scene, background clutter, and pedestrian appearance variation. Two-stage object detector has higher accuracy over one-stage detector due to foreground-background class imbalance. RetinaNet[14] is a one-stage object detector which solves the problem of extreme foreground-background class imbalance during training. This class imbalance is solved by reshaping the standard cross entropy loss in a way that it down-weights the loss assigned to well-classified samples. This reduces the impact of easy negatives in total loss and final loss mainly comes from a sparse set of hard examples. RetinaNet solves the speed vs accuracy problem of one-stage object detector and surpasses of all existing state-of-the-art two-stage detectors. Exploring RetinaNet with the focal loss in pedestrian detection problem can be a great idea because it is fast, accurate and one-stage.

2. Background And Related work

Pedestrian detection is already a well-studied topic. There are many papers and datasets showing significant improvement.

2.1. Feature-Based Methods

The classical way of pedestrian detection is supervised and unsupervised feature extraction and run a SVM classifier on top of this feature. Rowley[15] used neural networks for facial detection was a pioneer work of feature-based detection. Viola and Jones[20] made an important step towards pedestrian/object detection. They utilized an integral image representation and AdaBoost based method to detect object/pedestrian. Histograms of Oriented Gradients(HOG) adopted by Dalal and Triggs[19] for human and pedestrian detection. These were followed by improvements to the feature based methods in order to improve feature-based detection. These methods are feature-based detection along with SVM regression methods to perform pedestrian detection. survey papers by Dollar et al.[17,18] compares the different feature based methods in detail and evaluate their performance on the Caltech pedestrian database[16].

2.2. Convolutional Neural Network Based methods

Since 2012 after the introduction of AexNet, convoluted neural networks based methods have been state-of-the-art in various computer vision tasks including object detection. Deep Neural Nets like AlexNet[10] and GoogleNet[21] improved the ImageNet challenge quite a lot. Due to the success of deep neural networks in classification tasks, many researchers focus on detecting pedestrian and object using convolutional neural networks. Tome et al.[22] applied CNN to this problem. Fukui et al.[24], Sermanet et al.[25] and Angelova et al. [22] also contributed to pedestrian and object detection using the convolutional neural network. Above methods use a feature extraction fromthe image and then use CNN for classification. Molin, D [26] developed a method where the entire image as an input and in output it provides the probability of a pedestrian on the image pixels. R-CNN based region based convolutional neural network has been applied to pedestrian detection[27].

3. Methodologies

The Pedestrian detection methodology comprises of per-processing of the input image, data augmentation, and network model.

3.1. Pre-processing

We applied three pre-processing steps on the images. First, We normalized the images by subtracting the mean RGB value of ImageNet dataset as suggested in [10]. Then, The input pixel range was converted to 1 - 0. After that, All the images have been resized to the appropriate size to be fed into the neural network.

3.2. Data Augmentation

Data augmentation is exclusively used to generate more training instances and is essential for the improved generalization. It is a well-established practice for the training CNN. For the pedestrian detection task, each pedestrian in a training set is jittered in one of the following three ways. The first is randomly translating the pedestrian with 10% of the image size. The second is randomly rotating the image with an angle between −5 and 5 degree . The third is randomly scaling the pedestrian with a factor between 0.9 and These augmentation parameters will generate new training examples that are a little different from the original training data, but they still look like they are keeping the same label. Random flip, rotation were used as data augmentation method. Data augmentation is performed only on training data, and the augmented data are all added to the training set. we picked standards pedestrian templates, rotated, scaled and put them in an image with no pedestrian and with some noise like white noise and distortions.

3.3. Network Model

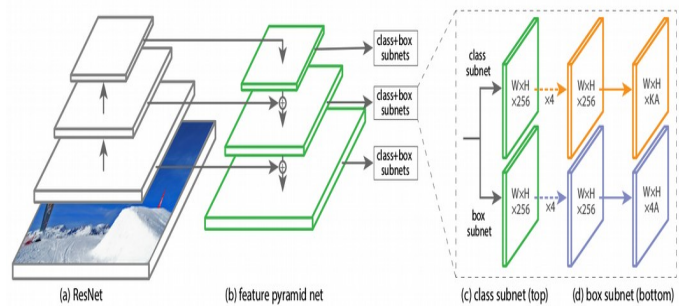

Fig. 1  The one-stage RetinaNet using Feature Pyramid Network (FPN) backbone on top of ResNet sub-networks for class prediction and bounding box prediction.

ResNet-152[14] was used as a feature extractor. Pyramid Network (FPN) is used [28] as the backbone network for RetinaNet. The classification subnet predicts the probability of object presence at each spatial position for each anchor and it's class. A small FCN to each pyramid level was used to regress the offset from each anchor box to a nearby ground-truth object.

4. Training Details

For training ResNet-152 was used to build the feature pyramid. The focal loss was used which is comprised of classification loss and regression loss. ResNet-152 was ImageNet per-trained. Adam optimizer was used with the learning rate of 0.001. Random flip, rotation were used as data augmentation method. We used a batch size of 4 to train the network. Our deep learning rig has 2 1080ti based system multi-GPU system. We effectively palatalized training procedures. Moreover, We employed batch size of 64, so that the maximum capacity of the GPUs can be utilized.

5. Testing

For testing purposes, We used 2 GPUs simulataneously to speed up the testing process. We selected coco evaluation paradigm of threshold of 0.5:.05:.95 to calculate the average precision. Again, After carefully testing for several times, We averaged the all average precision. Finally, Our proposed approach has obtained mAP 0.4061.

6. Conclusion

In this research, We have demonstrated that RetinaNet based neural networks achieved high-performance pedestrian detection in wider pedestrian detection challenge. In the future, We are looking forward to exploring more in the robustness of this pedestrian detection system in the other pedestrian detection datasets too. RetinaNet based focal loss can achieve competitive detection performance wider pedestrian detection dataset with minimum tuning of the hyper-parameters. Due to the varying sizes of the pedestrian, it is hard for the detector to detect the pedestrian in all resolution. Moreover, FPN largely solves this issue, still there is more scope for the improvement. In addition to that, a bigger dataset with improved feature variety will improve the overall performance of the detection system by learning better representation and reducing the risk of over-fitting. In addition to that, performing additional regularization tweaks and fine-tuning of hyper-parameters may improve the model's robustness. The proposed approach has mAP 0.4061.

References

[1] S. Ren, K. He, R. B. Girshick, and J. Sun. Faster R-CNN: towards real-time object detection with region proposal networks. CoRR, abs/1506.01497, 2015

[2] Liliang Zhang, Liang Lin, Xiaodan Liang, Kaiming He, Is Faster R-CNN Doing Well for Pedestrian Detection?, https://arxiv.org/abs/1607.07032

[3] M. Everingham, L. Gool, C. K. Williams, J. Winn, and A. Zisserman. The pascal visual object classes (voc) challenge. Int. J. Comput. Vision, 88(2):303–338, June 2010.

[1] J. Deng, W. Dong, R. Socher, L.-J. Li, K. Li, and L. Fei-Fei. ImageNet: A Large-Scale Hierarchical Image Database. In CVPR09, 2009.

[2] R. Girshick, J. Donahue, T. Darrell, and J. Malik. Rich feature hierarchies for accurate object detection and semantic segmentation. In Proceedings of the IEEE Conference on Computer Vision and Pattern Recognition (CVPR), 2014.

[3] R. B. Girshick. Fast R-CNN. CoRR, abs/1504.08083, 2015

[4] K. He, X. Zhang, S. Ren, and J. Sun. Spatial pyramid pooling in deep convolutional networks for visual recognition. IEEE Trans. Pattern Anal. Mach. Intell., 37(9):1904–1916, 2015.

[5] J. Hosang, R. Benenson, and B. Schiele. How good are detection proposals, really? In BMVC, 2014.

[6] J. Jin, K. Fu, and C. Zhang. Traffic sign recognition with hinge loss trained convolutional neural networks. IEEE Transactions on Intelligent Transportation Systems, 15(5):1991–2000, 2014.

[7] A. Krizhevsky, I. Sutskever, and G. E. Hinton. Imagenet classification with deep convolutional neural networks. In NIPS 2012

[8] T. Lin, M. Maire, S. Belongie, L. D. Bourdev, R. B. Girshick, J. Hays, P. Perona, D. Ramanan, P. Dollár, and C. L. Zitnick. Microsoft COCO: common objects in context. CoRR, abs/1405.0312, 2014.

[9] P. Sermanet and Y. LeCun. Traffic sign recognition with multi-scale convolutional networks. In Neural Networks (IJCNN), The 2011 International Joint Conference on, pages 2809–2813, July 2011.

[10] K. Simonyan and A. Zisserman. Very deep convolutional networks for large-scale image recognition. CoRR, abs/1409.1556, 2014 .

[11] Tsung-Yi Lin, Priya Goyal, Ross Girshick, Kaiming He, Piotr Dollár, Focal Loss for Dense Object Detection, https://arxiv.org/abs/1708.02002

[12] Rowley, A. H., Baluja, S., Kanade.,T, "Neural Network-Based Face Detection", IEEE Transactions on Pattern Analysis and Machine Intelligence, January 1998.

[13] http://www.vision.caltech.edu/Image_Datasets/CaltechPedestrians/

[14] Dollar,. P, Wojek, C., Schiele, B., Perona, P.,"Pedestrian Detection: An Evaluation of the State of the Art.", IEEE Transactions on Pattern Analysis and Machine Intelligence, ISSN:0162-8828, August 2011.

[15] Dollar, P., Wojek, C., Schiele, B., Perona, P.,"Pedestrian Detection: A Benchmark", IEEE Conference on Computer Vision and Pattern Recognition, 2009

[16] N. Dalal and B. Triggs, "Histograms of oriented gradients for human detection", IEEE Conference on Computer Vision and Pattern Recognition, 2005.

[17] Viola, P., Jones, M., "Robust Real-time Object Detection", 2nd International Workshop on Statistical and computational theories of vision, modeling, learning, computing, and sampling, 2001.

[18] Szegedy, C., Liu, W, Jia, Y., Sermanet, P., Anguelov, D., Erhan, D., Vanhoucke, V., Rabinovich, A.," Going Deeper with Convolutions", ILSVRC 2014

[19] Tome, D., Monti, F.,,. Baroffio, L., Bondi, L., Tagliasacchi, M., Tubaro, S., "Deep convoluted neural networks for pedestrian detection.", Preprint.Elsevier Journal of Signal Processing: Image Communication, 2015.

[20] Angelova, A., Krizhevsky, A., Vanhoucke, V., Ogale,A., Ferguson, D.,"Real-Time Pedestrian Detection with Deep Network Cascades", White Paper[10] Fukui, H., Yamashita, T., Yamauchi, Y., Fujiyoshi.,H,Murase, H., "Pedestrian Detection based on Deep Convolutional Neural Network with ensemble Inference Network", IEEE Intelligent Vehicles Symposium, 2015

[21] Fukui, H., Yamashita, T., Yamauchi, Y., Fujiyoshi., H, Murase, H., "Pedestrian Detection based on Deep Convolutional Neural Network with ensemble Inference Network", IEEE Intelligent Vehicles Symposium, 2015 25. Sermanet, P., Kavukcuoglu, K., Chintal

[22] Sermanet, P., Kavukcuoglu, K., Chintala, S., LeCun, Y. " Pedestrian Detection with unsupervised Multistage Feature Learning", International Conference on Computer Vision and Pattern Recognition, 2013.

[23] Molin, D., "Pedestrian Detection using convolutional neural networks", PhD Thesis, Department of Electrical Engineering, Linkopings Universitet, Sweden, 2015

[24] Li, J., Liang, X., Shen, S., Xu, T., Yan, S., "Scale- aware Fast R-CNN for Pedestrian Detection", arXiv preprint, arXiv:1510.08160 28. T.-Y. Lin, P. Dollár, R. Girshick, K.

He, B. Hariharan, and S. Belongie. Feature pyramid networks for object detection. In CVPR, 2017.